# Question Answering for Electronic Health Records: A Scoping Review of datasets and models

Jayetri Bardhan[a], Kirk Roberts[b,1], Daisy Zhe Wang[a]

[a]Department of Computer and Information Science and Engineering, University of Florida, Gainesville, FL, USA
[b]School of Biomedical Informatics, The University of Texas Health Science Center at Houston, TX, USA

**Abstract**
**Background:** Question Answering (QA) systems on patient-related data can assist both clinicians and patients. They can, for example, assist clinicians in decision-making and enable patients to have a better understanding of their medical history. Significant amounts of patient data are stored in Electronic Health Records (EHRs), making EHR QA an important research area. In EHR QA, the answer is obtained from the patient's medical record. Because of the differences in data format and modality, this differs greatly from other medical QA tasks that employ medical websites or scientific papers to retrieve answers, making it critical to research EHR question answering.
**Objective:** This study aimed to provide a methodological review of existing works on QA over EHRs. The objectives of this study were to (i) identify the existing EHR QA datasets and analyze them, (ii) study the state-of-the-art methodologies used in this task, (iii) compare the different evaluation metrics used by these state-of-the-art models, and finally (iv) elicit the various challenges and the ongoing issues in EHR QA.
**Methods:** We searched for articles from January 1st, 2005 to September 30th, 2023 in four digital sources including Google Scholar, ACL Anthology, ACM Digital Library, and PubMed to collect relevant publications on EHR QA. Our systematic screening process followed PRISMA guidelines. 4111 papers were identified for our study, and after screening based on our inclusion criteria, we obtained a total of 47 papers for further study. The selected studies were then classified into two non-mutually exclusive categories depending on their scope: 'EHR QA datasets' and 'EHR QA Models'.
**Results:** A systematic screening process obtained a total of 47 papers on EHR QA for final review. Out of the 47 papers, 25 papers were about EHR QA datasets, and 37 papers were about EHR QA models. It was observed that QA on EHRs is relatively new and unexplored. Most of the works are fairly recent. Also, it was observed that emrQA is by far the most popular EHR QA dataset, both in terms of citations and usage in other papers. We have classified the EHR QA datasets based on their modality, and we have inferred that MIMIC-III and the n2c2 datasets are the most popular EHR database/corpus used in EHR QA. Furthermore, we identified the different models used in EHR QA along with the evaluation metrics used for these models.
**Conclusions:** EHR QA research faces multiple challenges such as the limited availability of clinical annotations, concept normalization in EHR QA, as well as challenges faced in generating realistic EHR QA datasets. There are still many gaps in research that motivate further work. This study will assist future researchers in focusing on areas of EHR QA that have possible future research directions.

---

[1] Corresponding author. Address: School of Biomedical Informatics, The University of Texas Health Science Center at Houston, 7000 Fannin St #600, Houston, TX 77030, USA
*Email addresses:* jayetri.bardhan@ufl.edu (Jayetri Bardhan), kirk.roberts@uth.tmc.edu (Kirk Roberts), daisyw@cise.ufl.edu (Daisy Zhe Wang)

# Introduction

## Motivation

Medical QA may use biomedical journals, internet articles, as well as patient-specific data such as that stored in the Electronic Health Record (EHR) for QA. While there has been a lot of work in medical QA [1, 2, 3, 4, 5], most of these works do not help to answer patient-specific questions. In patient-specific QA, the answer is obtained from the patient's medical record (i.e., the EHR). This differs from other medical QA tasks due to linguistic issues (for example, EHR notes are very different in terminology, grammar, style, and structure from biomedical articles) and privacy limitations (for example, most biomedical articles have a publicly available abstract while there are laws in most countries limiting the sharing of patient records). Additionally, patient-specific QA also prevents the use of many common QA techniques (such as aggregating answers from different biomedical articles to give weight to a consensus opinion). All this merits the review of EHR QA separate from other medical QA approaches to properly scope its data and methods. In this review paper, our aim is to discuss all the recent approaches and methodologies used for question answering on electronic health records. There have been some reviews on medical question answering [6, 7], but none of the previous review papers have focused solely on EHR QA. To the best of our knowledge, this is the first work that does a scoping review of QA on EHRs and examines the various datasets and methodologies utilized in EHR QA. There are several aspects of EHR QA that merit analysis of scope.

One such aspect is data modality and the variety of methodological approaches available for EHR QA. The methodological approach used is determined by the format of the EHR data. EHRs contain structured data and unstructured data. Structured EHR data is based on standardized terminologies and ontologies and is often available in the form of relational databases. On the other hand, unstructured EHR data has minimal standardization and includes data types such as textual notes and clinical imaging studies. Two kinds of approaches are used for QA on structured

EHR data. In the first approach [5], the natural language questions are converted into structured queries (such as SQL). These queries are used to retrieve answers from the database. In the second approach [8], the structured EHR tables are converted into knowledge graphs, following which the natural language questions are converted into graph queries (such as SPARQL) in order to extract answers from the database. QA on unstructured clinical EHR notes is mostly performed as a reading comprehension task, where given a question and clinical notes as context, a span of text from the notes is returned as the answer. There can also be multimodal EHR QA which can use both structured and unstructured EHR data for QA. The aim of this study is to identify the studies that use EHR QA. We have further narrowed our search to EHR QA studies that use natural language processing (NLP) techniques on the questions, but may or may not use NLP on the answers. We have excluded studies in which questions are asked about images (e.g., radiology scans) as these questions and datasets have an entirely different focus.

The second aspect of EHR QA is the access to raw medical data. Due to privacy restrictions on clinical data, the replication and sharing of methods have been reduced compared to QA in other domains. This has led to the emphasis on sharable EHR datasets on which QA benchmarks can be made. MIMIC-IV [9] and the eICU Database [10] are large publicly available EHR databases for patients admitted to intensive care units. The MIMIC-III [11] database provides the foundation for a lot of the existing QA studies on EHRs. MIMIC-IV introduced in the year 2020, is a recent update to the MIMIC-III database. Finally, the n2c2 datasets (previously known as i2b2 datasets) is another repository of clinical notes, which have been used by the clinical QA community to develop EHR QA datasets.

Another aspect that warrants a scoping review of EHR QA is to study its different applications, including information extraction, cohort selection, and risk score calculation. For instance, Datta et al. (2020) [12] used a two-turn question answering approach to extract spatial relations from radiology reports. Similarly, Xiong et al. (2021) [13] used a QA approach with the help of a machine reading comprehension framework for cohort selection, where every selection criterion is converted into questions using simple rules. For example, the selection criteria "ALCOHOL-ABUSE" is converted to the question - "Current alcohol use over weekly recommended limits?". Following this, using state-of-the-art machine-reading comprehension models like BERT [14], BiDAF [15], BioBERT [16], NCBI-BERT, RoBERTa [17], and BIMPM [18], are used to match question and passage pairs in order to select cohorts. Furthermore, Liang et al. (2022) [19] demonstrates that QA over EHR data can improve risk score calculation.

Lastly, EHR QA systems face a variety of challenges ranging from parsing natural language questions to retrieving answers. In the case of structured data, the natural language question needs to be parsed and converted to a structured query which can be used to query the database. Medical terms from the queries, such as "blood pressure" and "leukemia," must be normalized into standard ontologies. Clinical text frequently uses acronyms for medical concepts. These abbreviations are often ambiguous (for example, "pt" can refer to the patient or physical therapy) [20] and so must be identified and standardized by the QA system before querying over the EHR database or clinical data. These problems are exacerbated by the fact that the standard NLP approaches to such issues require large amounts of labeled data from the domain of interest. Few such labeled EHR datasets exist. This is because annotating EHR QA datasets requires clinical expertise and is time-consuming. Existing general-domain QA systems provide erroneous results when they are not trained on clinical QA datasets. Additionally, the majority of the data found in EHRs is complex and contains both missing and inconsistent information [21, 22], which adds to the difficulty of performing QA on EHRs. In the Discussion section, we have provided more detailed explanations of the various challenges of using QA on EHRs.

The wide variety of challenges and barriers discussed above motivates the need for a systematic scoping review of EHR QA literature. This paper identifies the articles that fall under the scope of EHR QA, identifies the difficult challenges faced in the task, then enumerates both the data sources and QA methods that have been used to overcome such challenges. Finally, this paper also highlights the open issues in this field that demands future work in EHR QA.

**Template-based dataset generation**
Prior to diving into the methodology and results of this review, it is helpful to introduce a common semi-automated approach for building EHR QA datasets as essentially all large EHR QA datasets utilize this approach. This also impacts the screening process described in the next section. Generally speaking, large EHR QA datasets are often required to increase the performance of EHR QA models. However, the creation of these datasets necessitates subject expertise. The slot-filling approach to generate template-based datasets is a semi-automated process, and hence very popular. The majority of the EHR QA datasets are template-based [8, 5, 23, 24, 25]. The overall steps used to construct template-based QA datasets are illustrated using a flowchart in Figure 1.

To minimize the need for clinical experts' involvement in the dataset generation process, existing annotations of other non-QA clinical tasks (such as entity recognition and relations learning) are used for generating EHR question-answer pairs. The existing clinical annotations are used as proxy-expert in the dataset generation process [24]. In the first step, template questions containing placeholders (in place of entities) are constructed. An example of a question template is – "Has this patient ever been treated with |medication|?". Here, |medication|, |problem|, and |treatment| are some commonly *used* placeholders. These placeholders in the questions are then slot-filled to obtain QA pairs using the entities in the EHR data and database schema (for structured EHR database) with the help of the existing annotations from the clinical NLP datasets. So, in a question template such as - "Has this patient ever been treated with |medication|?", entities like "insulin" and "tylenol" from the EHR database/clinical notes (sharing the same entity type as |medication|) are slot-filled in the question template to obtain questions such as - "Has this patient ever been treated with insulin?" and "Has this patient ever been treated with tylenol?". Following this approach, the RxWhyQA [25] and DrugEHRQA [23] datasets use the existing annotations from the 2018 National NLP Clinical Challenges (n2c2) corpus, and the emrQA and emrKBQA datasets use annotations from six clinical tasks from the n2c2 repository [26, 27, 28, 29, 30, 31]

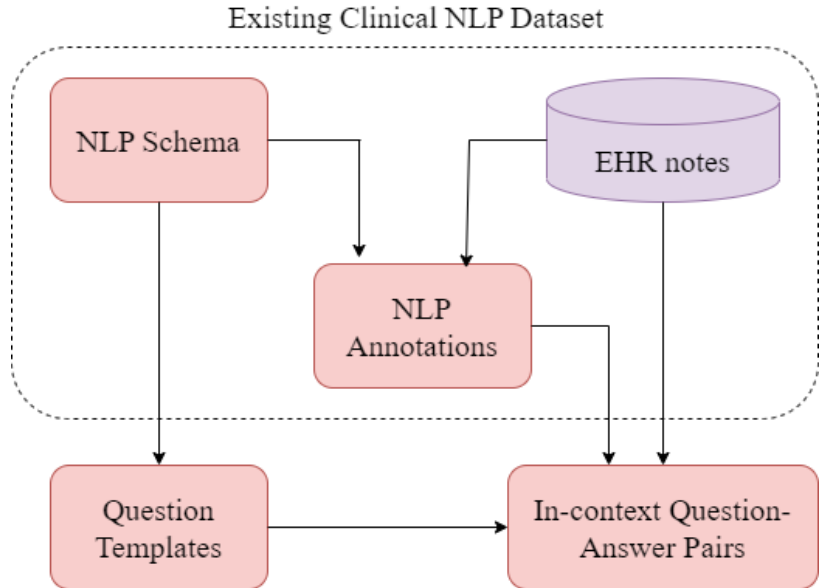

Figure 1. Flowchart for template-based dataset generation process.

While generating the emrQA and emrKBQA dataset, logical form templates (which is used to map EHR schema or ontology to represent relations in the questions) are annotated by clinical experts for different question templates. If more than one question templates map to the same logical form template, then they are considered paraphrases of each other. In the emrQA dataset, clinical expert annotations of non-QA tasks such as entity recognition, relation learning, coreference, and medication challenge annotations (in the n2c2 repository) were used to slot-fill placeholders in question and logical form templates, which in turn were used to generate answers. For example, the medication challenge in the n2c2 repository has annotations for medication and their corresponding dosage (for example, medication=Nitroglycerin, dosage=40mg). This was used to generate instances of the question "What is the dosage of |medication|?", along with instances of its corresponding logical form MedicationEvent(|medication|)[dosage=x]. The dosage value, i.e. 40mg is the answer to the question. Similarly, the heart disease challenge dataset contains temporal information, was used to derive temporal-reasoning related question-answer pairs. The emrKBQA dataset used the same question templates and logical form templates of emrQA, which were then slot-filled using entities from the MIMIC-III KB [11]. The answers of the emrKBQA dataset are present in the table cells of the MIMIC-III KB. The entity types used in the placeholders are test, problem, treatment, medication, and mode. So far, the slot-filling QA dataset generation process has proven to be the most common method of generating EHR QA datasets. This is because, while some manual annotation from domain experts is necessary, most of the process is automated.

## Methods

### Search Process

This study aims to review existing research on question answering over electronic health records. This includes papers on EHR QA datasets, QA models, and various approaches proposed over the years. We included papers related to question answering in the clinical domain, specifically in EHRs. Papers in which EHRs are not used have been excluded. We have fulfilled all PRISMA scoping review requirements and have attached a completed copy of PRISMA checklist in Multimedia Appendix 1. The flowchart for conducting this study is shown in Figure 2.

Each of the data sources has been queried to search for papers with the title having at least one of the following keywords: "clinical", "medical", "patient", "EHR", EMR", "Electronic Health Record(s)", "Electronic Medical Record(s)", or "patient". This should be used in a combination with one or more of the keywords: "question answering", "questions", "text to SQL", "reading comprehension", "machine comprehension", "machine reading", or "queries". The search was limited to the period from January 1st, 2005 to September 30th, 2023, so as to review only recent works. In this record identification and collection step, a total of 4285 papers were collected (i.e. 2790 from Google Scholar, 114 papers from ACM Digital Library, 72 papers from ACL Anthology, and 1309 papers from PubMed). Following this, we removed the duplicate papers and obtained a total of 4111 papers.

### Screening Process

We used a two-step screening process. The first step of screening involved an abstract and title screening process. We read the abstract and titles of all the papers and included papers only that were about EHR question answering. In this step, we obtained a total of 126 papers. In this step, we removed many irrelevant papers which just focused on "clinical questions" and "patient questions" but did not use NLP. We also removed non-research articles (such as PhD dissertations and books).

In the final stage of screening, a full-text review was used to screen the papers further. Papers that were about query engines and tools, and which did not use natural language questions, were removed. We excluded papers in languages other than English. We also removed papers that just had an abstract and did not contain full text. There were some papers that were about information retrieval systems, and not specifically QA. These were also excluded. Furthermore, we have excluded studies in which questions are asked about images or ECG as these studies have an entirely different focus. At the end of this step, we obtained a total of 37 papers. After the two-stage screening process, we performed forward snowballing, adding ten more papers after full-text review that cited the 37 previously included papers according to Google Scholar. Finally, we obtained a total of 47 papers at the end of our study. Then, we conducted an in-depth review of the final 47 papers which are discussed in the Results section.

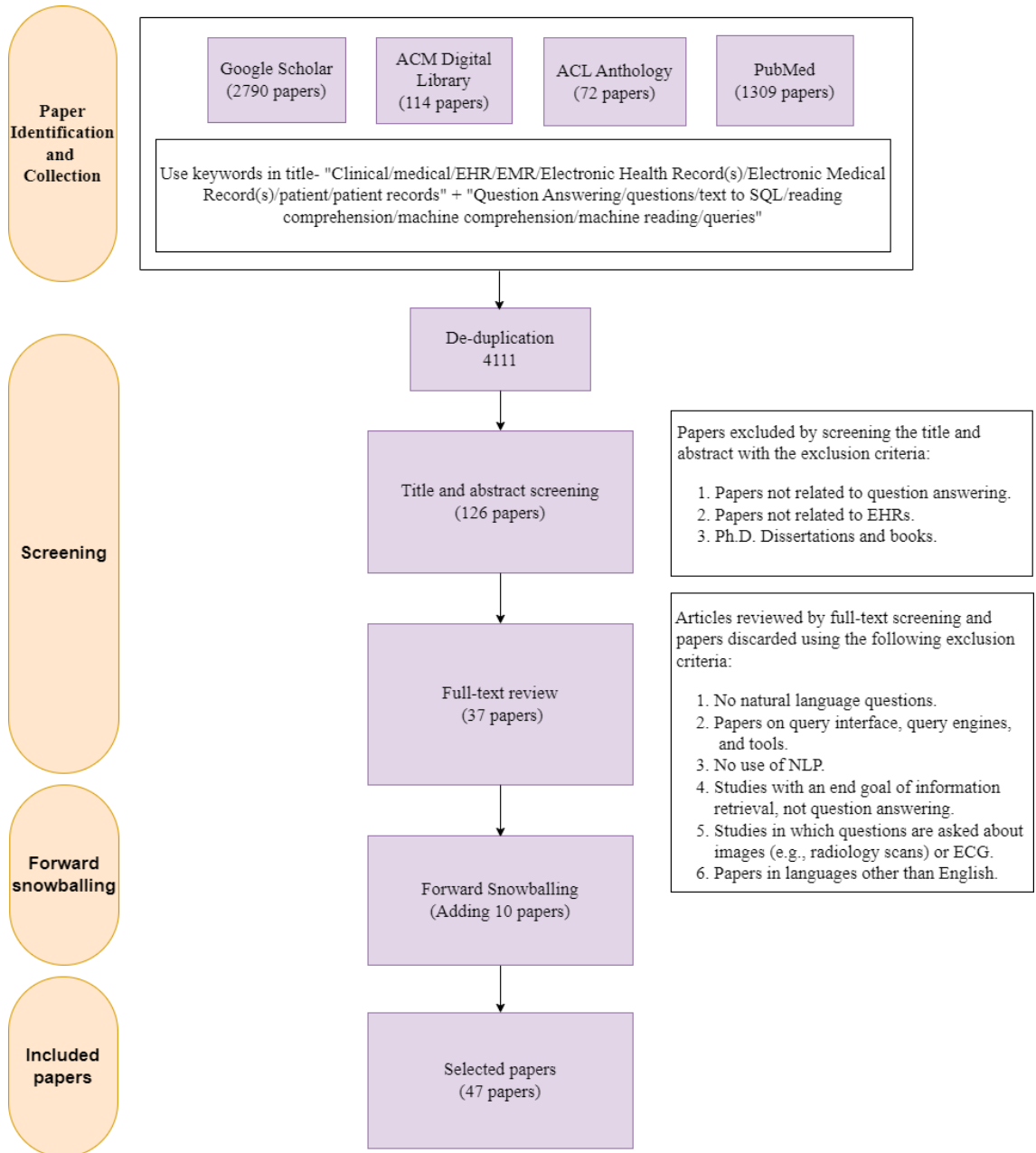

Figure 2. PRISMA diagram for study on QA over EHRs.

For this study, all the authors (JB, KR, DZW) jointly made the rules for inclusion and exclusion criteria, that were used during paper collection and screening process. Based on the rules decided, JB collected the papers and worked on the overall screening process. For papers that were borderline for inclusion were independently screened by KR and then were resolved after discussion. The final list made during the full-text review process was again independently screened and reviewed by JB and KR, with conflicts resolved after discussion.

## Results

### Classification of Selected Papers

This section presents the findings of our study about existing EHR QA papers.
Table *1* lists our final list of selected publications post-screening and then classified the papers based on their scope - 'EHR QA Datasets' and 'EHR QA Models'. We have further classified the studies on EHR QA models based on their function in the QA pipeline. 'Full QA' denotes the papers on EHR QA models which are about end-to-end EHR QA systems. In the remaining part of the paper, we have provided our in-depth analysis of studies on QA using EHRs. In Multimedia Appendix 2, we have summarized our final list of selected papers.

Table 1. List of included papers in the systematic review

| Type of study | References | |
|---|---|---|
| EHR QA datasets | [5, 8, 23, 24, 25, 32, 33] [34, 35, 36, 37, 38, 39, 40] [41, 42, 43, 44, 45, 46, 47] [48, 49, 50, 51] | |
| **EHR QA models** | | |
| | Question generation | [43] |
| | Question paraphrasing | [52, 53, 54] |
| | Question classification | [55, 56] |
| | Full QA | [5, 8, 23, 24, 25, 38, 39, 40, 41] [42, 48, 49, 50, 51, 57, 58, 59] [60, 61, 62, 63, 64, 65, 66, 67] [68, 69, 70, 71, 72, 73] |

Figure 3 illustrates the number of publications on EHR QA over the years. From Figure 3, it can be observed that this is a relatively very new field and most of the publications in this domain are fairly recent. Note that since this systematic review is conducted based on studies published before September 30th, 2023, hence the number of studies shown for the year 2023 is recorded only for a period of 9 months. In the following subsections, we discuss our findings on existing EHR QA datasets, the various models used for questioning over EHRs; and also, the different evaluation metrics used.

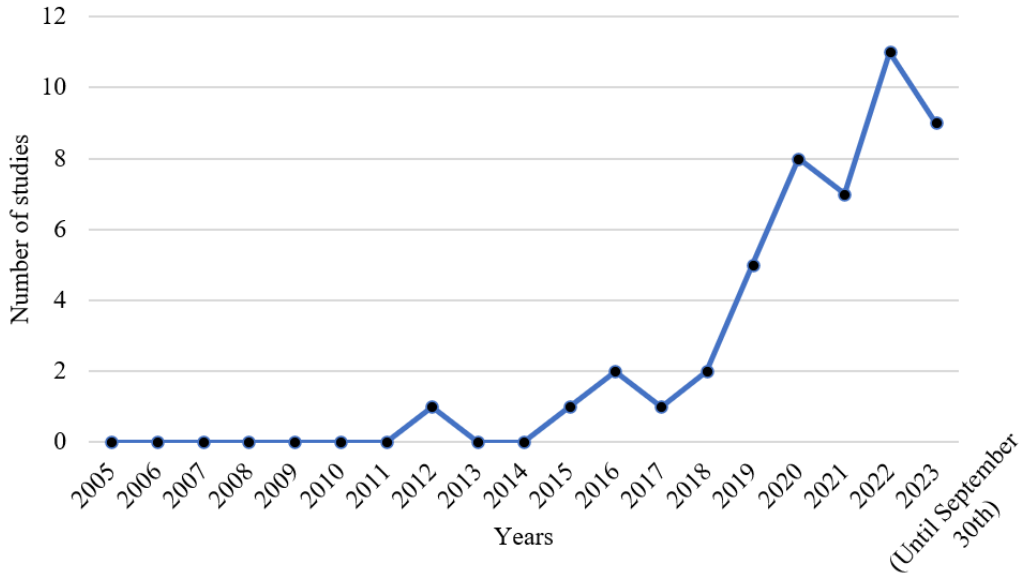

Figure 3. Number of studies on EHR QA over the years.

**Datasets**

*Dataset Classification and Analysis*
Table 2 displays the total number of citations of all the EHR QA datasets (as noted on September 30th, 2023). It also lists the number of studies included in our review that have used these datasets. We can observe from the figures that emrQA [24] is the most popular out of all the other EHR QA datasets. This is likely due to emrQA's size (1,295,814 question-logical forms and 455,837 question-answer pairs) and similarity to the SQuAD-QA format.

Table 2. Popularity of EHR QA datasets

| Datasets | No. of citations | No. of studies on EHR QA using the datasets |
|---|---|---|
| emrQA [24] | 151 | 11 |
| MIMICSQL [5] | 51 | 3 |
| Yue et al. (2020) [46] | 40 | 0 |
| MIMICSPARQL* [41] | 27 | 2 |
| Yue et al. (2021) [42] | 18 | 0 |
| Roberts et al. (2016) [32] | 18 | 3 |
| emrKBQA [8] | 15 | 0 |
| Raghavan et al. (2018) [34] | 13 | 0 |
| Roberts et al. (2015) [33] | 10 | 1 |
| Soni et al. (2019) [44] | 7 | 3 |
| Fan et al. (2019) [35] | 7 | 0 |
| DrugEHRQA [23] | 5 | 0 |
| DiSCQ [43] | 6 | 0 |
| Oliveira et al. (2021) [38] | 3 | 0 |
| RadQA [37] | 3 | 1 |
| EHRSQL [36] | 3 | 0 |
| Kim et al. (2022) [39] | 2 | 0 |
| ClinicalKBQA [40] | 2 | 0 |
| Hamidi et al. (2023) [48] | 1 | 0 |
| MedAlign [49] | 1 | 0 |
| RxWhyQA [25] | 0 | 0 |
| Mishra et al. (2021) [45] | 0 | 0 |
| CLIFT [47] | 0 | 0 |
| Mahbub et al. (2023) [50] | 0 | 0 |
| Dada et al. (2023) [51] | 0 | 0 |

The classification of EHR QA datasets is shown in Figure 4. EHR QA datasets can be unimodal or multimodal. Unimodal EHR QA datasets are based on question answering over one modality which can be in the form of structured EHR data or unstructured EHR clinical notes. Multimodal EHR QA datasets use both modalities for QA over EHRs. The DrugEHRQA [23] and MedAlign [49] datasets are example of multimodal EHR QA datasets that uses structured and unstructured EHR data for QA. Figure 5 shows the size and modalities of the different EHR QA datasets.

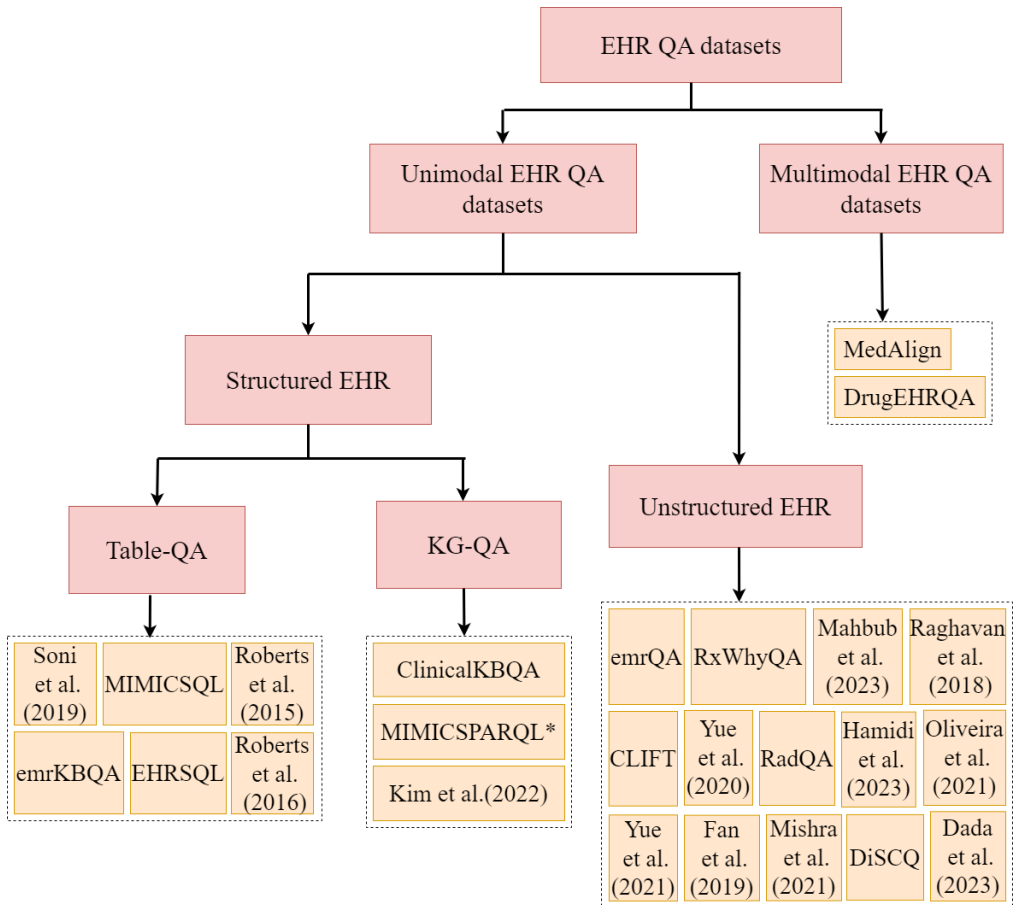

Figure 4. Classification of EHR QA datasets based on modality.

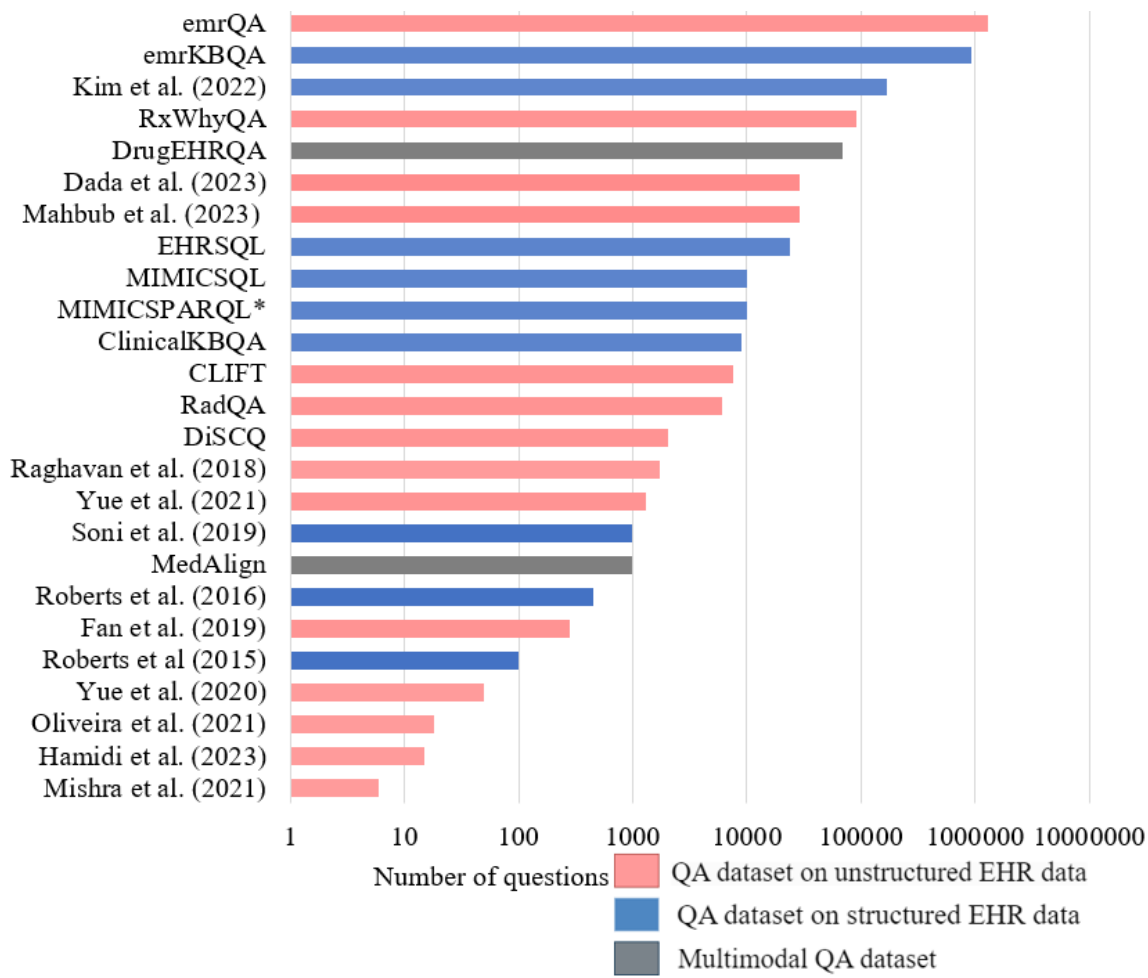

Figure 5. Plot for number of questions of different EHR QA datasets.

It is to be noted that the dataset introduced in Mishra et al. (2021) [45] uses six key questions (as can be observed from Figure 5), i.e. the same six questions have been re-used for all the articles. Multimedia Appendix 3 summarizes the existing EHR QA datasets. It should be noted that in Multimedia Appendix 3, "Database/Corpus" refers to the EHR database or clinical annotations from which the QA datasets are based on. These EHR databases or corpora contain answers to the questions. From the table in Multimedia Appendix 3, we can infer that most of the EHR QA datasets on structured EHR data use the MIMIC-III database [5, 8, 36, 39, 41], while most of the QA datasets on unstructured data use the n2c2 repository [24, 25, 35] or the clinical notes of MIMIC-III [37, 42, 43, 45, 46, 47, 48].

The following sections describe the QA datasets based on unimodal (structured or unstructured) and multimodal EHR data in detail.

*QA datasets based on unstructured EHR data*

Unstructured free text EHR data comprises discharge summaries, radiology reports, lab reports, medical images, progress notes, and many more note types. It accounts for roughly 80% of all EHR data [74]. One way to make use of this is to create a QA system that can extract answers from unstructured EHR data. Most of the QA datasets on unstructured clinical data are designed for the task of machine comprehension. Given clinical notes (containing patient information) and natural language questions, the objective of these tasks is to retrieve a span of text from the clinical notes as the answer.

emrQA [24], the most popular among the EHR QA datasets, contains 455,837 question-answer samples along with 1,295,814 question-logical form pairs. It relies on expert annotated n2c2 datasets [26, 27, 28, 29, 30, 31]. A semi-automatic template-based process was used to generate the dataset. From Figure 5, we can observe that the emrQA is the largest EHR QA dataset overall.

In spite of emrQA's popularity, it has some flaws. The emrQA dataset has attempted to simulate clinicians' questions using pre-defined templates and generating QA datasets by slot-filling with entities. As a result, the questions in the emrQA dataset are not very realistic or relevant to the medical community. They are also highly repetitive. For example, it was shown in Yue et al. (2020) [46] that the same model performance was obtained by sampling 5-20% of the dataset as with the entire dataset. This makes it necessary to create datasets that are more realistic and closer to real physicians' questions. Later, Yue et al. (2021) [42] developed 975 human-verified questions along with 312 human-generated questions based on 36 discharge summaries from MIMIC-III's clinical notes. After randomly sampling 100 questions individually the 975 human-verified questions and 312 human-generated questions, it was learned that 96% of the [42]'s human-verified questions were obtained from the emrQA's templates, and 54% of the human-generated questions of Yue et al. (2021) [42] used the same templates from emrQA.

The RxWhyQA dataset [25] and Fan et al. (2019) dataset [35] have reasoning-based questions. The RxWhyQA dataset contains a combination of reasoning-based unanswerable and multi-answer questions. Similar to the emrQA dataset, RxWhyQA is also a template-based dataset, and hence not very realistic. This made it necessary to create datasets that are more realistic and closer to real physicians' questions. The DiSCQ dataset [43] was created to address this issue and included questions about clinically relevant problems by gathering questions that clinicians could ask. It includes 2029 questions and over 1000 triggers based on MIMIC-III discharge reports. Each question in the DiSCQ dataset is paired with a segment of text that prompted the query. While annotating the DiSCQ dataset, the annotators read the discharge summaries and made a note of all questions that might be useful to the patient, while also recording the span of text that prompted the question.

The majority of the QA on unstructured EHR datasets are based on discharge summaries [24, 25, 35, 43, 45, 75]. RadQA [37] and Dada et al. (2023) [51] are the only two QA dataset that uses radiology reports for question answering. The types of questions used in the EHR QA datasets vary greatly from one another. emrQA covers different types of questions including Factual ("What"/"Show me"), reasoning ("How"/"Why"), and class prediction ("Is"/"has"). But the distribution of questions for the emrQA dataset is skewed, i.e., a majority of the questions in the emrQA dataset start with "what". In comparison, the authors of RadQA claim that the questions in their dataset are more evenly distributed than emrQA. The RxWhyQA dataset [25] and Fan et al.(2019) [35] are reasoning-based questions and hence their questions have "why-cues". Raghavan et al. (2018) [34] predominantly has temporal questions along with questions on presence/absence (i.e. "yes" or "no" questions) as well as questions on medications, tests, and procedures. Mishra et al. [45] on the other hand restricts itself to diagnosis-related questions. Table 3 compares some of the EHR QA datasets using unstructured EHR data for QA. Note that in Table 3, the length of questions and articles are expressed in terms of the number of tokens. Out of the fourteen QA datasets on unstructured EHR notes, only four of them (RadQA [37], RxWhyQA [25], Hamidi et al. (2023) [48], and Dada et al. (2023) [51]) contain unanswered questions.

Table 3. Comparison of different EHR QA datasets on unstructured data.

| Dataset | Database/ Corpus | Mode of dataset generation | Total questions | Unanswered questions | Average question length | Total articles | Average article length |
|---|---|---|---|---|---|---|---|
| emrQA [24] | n2c2 annotations (mostly discharge summaries) | Semi-automatically generated | 1,295,814 | 0 | 8.6 | 2425 | 3825 |
| RxWhyQA [25] | MIMIC-III (discharge summaries) | Automatically derived from the n2c2 2018 ADEs NLP challenge | 96,939 | 46,278 | - | 505 | - |
| Raghavan et al. (2018) [34] | Cleveland Clinic (medical records) | Human-generated (Medical students) | 1747 | 0 | - | 71 | - |
| Fan et al. (2019) [35] | 2010 n2c2/VA NLP challenge (discharge summaries) | Human-generated (Author) | 245 | 0 | - | 138 | - |
| RadQA [37] | MIMIC-III (radiology reports) | Human-generated (Physicians) | 6148 | 1754 | 8.56 | 1009 | 274.49 |
| Oliveira et al. (2021) [38] | SemClinBr corpus | Human-generated (Author) | 18 | 0 | - | 9 | - |

| Reference | Dataset | Generation Method | # Questions | # Unanswerable | Avg Q Length | # Articles | Avg Article Length |
|---|---|---|---|---|---|---|---|
| | (Portuguese nursing and medical notes) | | | | | | |
| Yue et al. (2021) [42, 75] | MIMIC-III (Clinical notes) | Trained question generation model paired with a human-in-the-loop | 1287 | 0 | 8.7 | 36 | 2644 |
| DiSCQ [43] | MIMIC-III (discharge summaries) | Human-generated (medical experts) | 2029 | 0 | 4.4 | 114 | 1481 |
| Mishra et al. (2021) [45] | MIMIC-III (discharge summaries) | Semi-automatically generated | 6 questions/article | - | - | 568 | - |
| Yue et al. (2020) [46] | MIMIC-III (Clinical notes) | Human-generated (medical experts) | 50 | 0 | - | - | - |
| CLIFT [47] | MIMIC-III | Validated by human experts | 7500 | 0 | 6.42, 8.31, 7.61, 7.19, and 8.40 for Smoke, Heart, Medication, Obesity, and Cancer datasets | - | 217.33, 234.18, 215.49, 212.88, and 210.16 for Smoke, Heart, Medication, Obesity, and Cancer datasets respectively |
| Hamidi et al. (2023) [48] | MIMIC-III (Clinical notes) | Human-generated | 15 | 5 | - | - | - |
| Mahbub et al. (2023) [50] | VA Corporate Data Warehouse (CDW) (Clinical progress notes) | Combination of manual exploration and rule-based NLP methods | 28855 | - | 6.22 | 2336 | 1003.98 |
| Dada et al. (2023) [51] | Radiology reports related to brain CT scans | Human-generated (medical student assistants) | 29,273 | - (Yes) | - | 1223 | - |

*QA datasets based on structured EHR data*

EHR tables contain patient information such as diagnoses, medications prescribed, treatments, procedures recommended, lab results details, and so on. It also includes a lot of temporal information, such as the date of admission, the date of discharge, and the duration of certain medications. The goal of QA tasks over structured databases is to translate the user's natural language question into a form that can be used to query the database.

The QA task on structured EHRs can be classified into two types based on the two most common forms of structured data: relational databases and knowledge graphs. The first type of QA task entails converting natural language questions into SQL (or logical form) queries that can be used to query the database. In the other type of approach, the EHR data exists in the form of knowledge graphs containing patient information, and the natural language questions are often converted into SPARQL queries to retrieve the answer. MIMICSQL, emrKBQA, and EHRSQL are examples of datasets that use table-based QA approach, whereas datasets like ClinicalKBQA and MIMIC-SPARQL* use KG-based QA approach.

MIMICSQL [5] is a large dataset used for question-to-SQL query generation tasks in the clinical domain. The MIMICSQL dataset is based on the tables of the MIMIC-III database. emrKBQA [8] is the counterpart of the emrQA dataset, for question answering on structured EHRs. It is the largest QA dataset on structured EHR data (shown in Figure 5) and contains 940,000 samples of questions, logical forms, and answers. EHRSQL [36] is a text-to-SQL dataset for two publicly available EHR databases - MIMIC-III [11] and eICU [10]. It is the only QA dataset on structured EHR data that contains unanswerable questions. Other QA datasets for structured EHR databases include MIMIC-SPARQL* [41] and ClinicalKBQA [40]. However, unlike previous table-based QA datasets, these are knowledge graph-based QA datasets.

The MIMICSQL dataset [5] was created by making changes to the MIMIC-III database's original schema. Nine tables from the MIMIC-III database were merged into five tables in order to simplify the data structure. The derived tables and schemas were not the same as those found in actual hospitals and nursing homes. As a result, a model trained on the MIMICSQL dataset will not be able to generalize to a real-world hospital setting. To address this issue, Park et al. (2021) [41] introduced two new datasets - a graph-based EHR QA dataset (MIMIC-SPARQL*) and a table-based EHRQA dataset (MIMICSQL*). This was done to improve the analysis of EHR QA systems and to investigate the performance of each of these datasets. MIMICSQL [5] was modified to create MIMICSQL* in order to comply with the original MIMIC-III database schema [11]. The graph counterpart of the MIMICSQL* dataset is MIMIC-SPARQL*. Figure 6 compares the two datasets - MIMICSQL and MIMICSPARQL* based on the length of the questions and the length of SQL/SPARQL queries.

Wang et al. (2021) [40] generated a clinical knowledge graph (ClinicalKB) with the help of clinical notes of n2c2 annotations and linked different patient information in order to perform knowledge base QA. The ClinicalKB links clinical notes of patients, allowing questions about different patients to be answered. At the same time, ClinicalKB contains clinical notes which allow questions not present in the database.

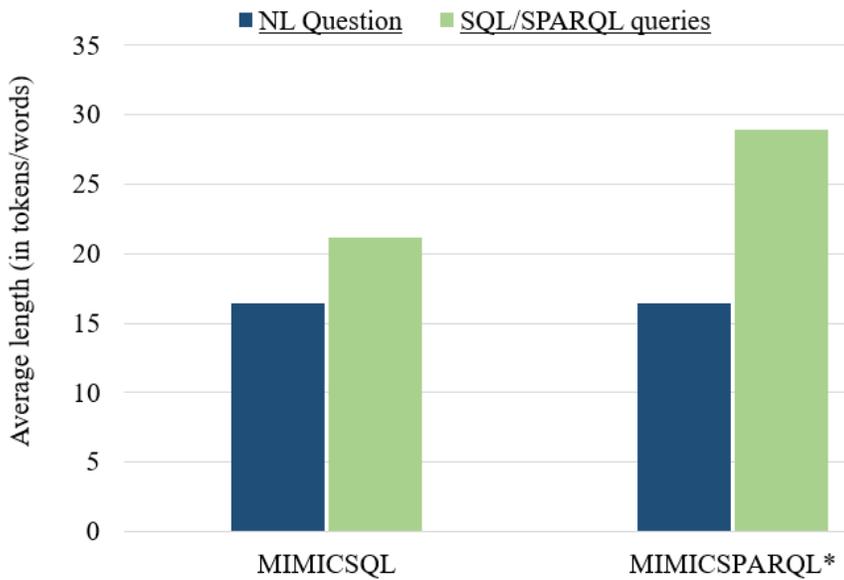

Figure 6. Average length of questions and SQL/SPARQL queries (in tokens or words) for MIMICSQL and MIMICSPARQL* datasets.

Furthermore, Wang et al. (2021) [40] generated the ClinicalKBQA dataset that can answer statistics-related questions about different patients as well as questions specific to individual patient records.

Roberts et al. [32, 33] and Soni et al. (2019) [44] introduced datasets where logical forms (based on lambda calculus expressions) were created for questions in order to perform QA on EHR data (known as semantic parsing). Soni et al. (2019) [44] constructed the question-logical form dataset with the help of Fast Healthcare Interoperability Resources (FHIR) server. Roberts et al. (2015) [33] annotated logical forms for 100 EHR questions, and Roberts et al. (2016) [32] extended this to 446 patient-related questions.

### *QA datasets based on multimodal EHR data*
Multimodal question answering is QA over more than one modality. QA over more than one modality can help in seeking more accurate answers while taking advantage of more than one source for QA. DrugEHRQA [23] is the first multimodal EHR QA dataset. It uses both structured tables of MIMIC-III and unstructured clinical notes for QA. The DrugEHRQA dataset is a template-based dataset containing medicine-related queries, its corresponding SQL queries for querying over multi-relational EHR tables, the retrieved answer from one or both modalities, as well as the final multi-modal answer. The MedAlign dataset [49] also utilizes structured and unstructured EHR data for QA, but indirectly. The instructions and response pairs of the MedAlign dataset are based on XML markup documents that are derived from structured and unstructured EHR data.

**Models and Approaches for QA on EHRs**

This section describes the various QA models used in EHRs. QA tasks vary depending on the EHR modality since different information is found in different modalities. Most QA models on clinical notes use a machine reading comprehension approach (MRC), i.e. for a given question, the QA model is trained to predict the span of text containing the answer from the clinical note [24, 25, 38, 42, 48, 50, 51, 59, 65, 66]. For question answering over EHR tables, translating questions to SQL queries is one of the major approaches used to retrieve answers from the EHR tables [5, 61, 62, 71]. The other approach is to transform the EHR relational database into a knowledge graph and perform a knowledge-graph QA task [39, 41, 61]. Table 4 summarizes all the QA models (Full QA) used for EHRs.

Table 4. Summary of models for QA over EHRs.

| Papers | Task | Answer Type | Dataset | Model |
|---|---|---|---|---|
| Pampari et al. (2018) [24] | MRC | Text span | emrQA | For question answering task: DrQA's document reader and a multi-class logistic regression model for predicting class.<br><br>For question-to-logical form task: A sequence-to-sequence model is used with attention paradigm. |
| Moon et al. (2023) [25] | MRC | Text span | RxWhyQA | Clinical BERT model with incremental masking to generate multiple answers. |
| Oliveira et al. (2021) [38] | MRC | Text span | SQUAD dataset [76] in Portuguese and another QA dataset developed in Portuguese | Used BioBERTpt (A deep contextual embedding model to support Portuguese language for biomedical data). |
| | | | from SemClinBr corpus | |
| Yue et al. (2021) [42] | MRC | Text span | emrQA and [42, 75] as test set. | For question answering task: DrQA's DocReader and ClinicalBERT are used.<br><br>For question generation task: Question Phrase Prediction (QPP) module is used along with the base question generation models (NQG, NQG++, and BERT-SQG). |
| Hamidi et al. (2023) [48] | MRC | Text span | QA dataset constructed based on TREC 2016 Clinical Decision Support Track [77] | ChatGPT (versions 3.5 and 4), Google Bard, and Claude |
| Fleming et al. (2023) [49] | Multi-step refinement approach using standard prompt template | Response based on XML markup derived from EHR data | MedAlign dataset | Six language models – GPT-4 (32K tokens+ multi-step refinement), GPT-4 (32K tokens), GPT4 (2K tokens), Vicuña-13B (2K tokens), Vicuña-7B (2K tokens), and Vicuña-7B (2K tokens) |
| Mahbub et et al. (2023) [50] | MRC | Text span | QA dataset on Injection Drug use constructed | Baseline models: Four state-of-the-art pre-trained language models - BERT, BioBERT, BlueBERT, and ClinicalBERT used for QA.<br><br>Modeling with transfer learning: Sequential learning and Adversarial learning |

| Study | Task | Answer Type | Dataset | Method |
|---|---|---|---|---|
| Dada et al. (2023) [51] | MRC | Text span | Reading comprehension question answering constructed based on radiology report | G-BERT and GM-BERT (G-BERT further pre-trained on German medical articles) |
| Roberts et al. (2017) [57] | question to logical form | Text span | Annotation of 446 questions in [32] | Hybrid semantic parsing method, uses rule-based methods along with a machine learning-based classifier. |
| Rawat et al. (2019) [59] | MRC | Text span | Naranjo Scale Questionnaire | Employs multi-level attention layers along with local and global context while answering questions. |
| Rawat et al. (2020) [60] | MRC | Text span | emrQA and MADE-QA dataset | Multitask learning with BERT and ERNIE [78] as the base model. |
| Wen et al. (2020) [64] | MRC | Text span | n2c2 notes, emrQA$_{why}$, and SQuAD$_{why}$ | BERT model trained on different data sources. |
| Soni et al. (2020) [65] | MRC | Text span | CliCR [79] and emrQA dataset | BERT, BioBERT, Clinical BERT, XLNet |
| Mairittha et al. (2020) [66] | MRC | Text span | why-question answering (why-QA) dataset developed based on 2010 n2c2/VA Workshop on Natural Language Processing Challenges for Clinical Records | Makes use of BERT (Large, Uncased, Whole Word Masking), BERT fine-tuned on Stanford Question Answering Dataset (SQuAD) benchmark, BioBERT, and an extended BioBERT fine-tuned on unstructured EHR data |

| Moon et al. (2022) [67] | MRC | Text span | Why-QAs from the n2c2 ADE Challenge and Medication Why-QAs from the emrQA | ClinicalBERT model fine-tuned on SQuAD-why dataset |
|---|---|---|---|---|
| Li et al. (2023) [68] | MRC | Text span | emrQA | Clinical-Longformer and Clinical-BigBird language model |
| Yang et al. (2022) [69] | MRC | Text span | emrQA | GatorTron language model |
| Lehman et al. (2023) [73] | MRC | Text span | RadQA | Evaluates 12 different language models ( T5-Base, Clinical-T5-Base-Ckpt, Clinical-T5-Base, RoBERTa-Large, BioClinRoBERTa, GatorTron, T5-Large, Clinical-T5-Large, PubMedGPT, T5-XL, Flan-T5-XXL, GPT-3) ranging from 220M to 175B parameters on three tasks including MRC task on EHR QA dataset. |
| Kang and Baek et al. (2022) [70] | Knowledge conditioned Feature Modulation on Transformer for MRC | Text span | emrQA | Knowledge-Augmented Language model Adaptation (KALA) |
| Wang et al. (2020) [5] | question to SQL query | Table content | MIMICSQL | TRanslate-Edit Model for Question-to-SQL (TREQS) |
| Raghavan et al. (2021) [8] | question to logical forms | Table content | emrKBQA | Min et al. (2020) [80] for sequence-to-sequence task along with ParaGen and ParaDetect model. |

| Pan et al. (2021) [62] | question to SQL query | Table content | MIMICSQL | Medical text–to-SQL model (MedTS) model |
|---|---|---|---|---|
| Soni et al. (2022) [63] | question to logical forms | Table content | ICU$_{data}$ [32] and FHIR$_{data}$ [44] | Tranx, Coarse2Fine, Transformer, Lexicon-based. |
| Tarbell et al. (2023) [71] | question to SQL query | Table content | MIMICSQL 2.0 split | T5 language model for question-to-SQL task, along with data augmentation method for back-translation. Then, the model is trained on both the MIMICSQL and Spider datasets to improve generalizability. |
| quEHRy [72] | question to answer extraction pipeline | Table content | FHIR$_{data}$ [44] and ICU$_{data}$ [32] | End-to-end EHR QA pipeline with concept normalization (MetaMap), time frame classification, semantic parsing, visualization with question understanding, and query module for FHIR mapping/processing |
| Kim et al. (2022) [39] | question to Program | Element from knowledge graph | MIMICSPARQL* | Program-based model |
| Wang et al. (2021) [40] | KBQA | Element from knowledge graph | ClinicalKBQA | Attention-based aspect reasoning (AAR) method for KBQA |
| Park et al. (2021) [41] | question to SPARQL query | Element from knowledge graph | MIMICSPARQL* | Seq2Seq model [81] and TREQS [5] |
| Schwertner et al. (2019) [58] | question to SPARQL query | Element from knowledge graph | QA dataset developed on Oncology XML EHR notes in Portuguese | Developed a framework ENSEPRO, which supported question answering |

| | | | | from Knowledge Bases. |
|---|---|---|---|---|
| Bae et al. (2021) [61] | question to query (SQL/ SPARQL) | Table content or element from knowledge graph | MIMICSQL* and MIMICSPARQL* | Unified encoder-decoder architecture that uses input masking (UniQA) |
| Bardhan et al. (2022) [23] | Multimodal QA | Text span or Table content | DrugEHRQA | MultimodalEHRQA |

We can observe from Table 4 that over the years, DrQA's document reader, BERT and ClinicalBERT are some of the most popular QA models used for unstructured clinical notes [24, 25, 42, 50, 60, 64, 65, 66, 67]. But since the year 2022 there has been a sharp rise in the number of studies introducing new large language models (besides BERT and other variants of BERT) for machine reading comprehension tasks [48, 68, 69, 73]. For example, Clinical-Longformer and Clinical-BigBird [68] and GatorTron [69] language models were proposed for various tasks including EHR QA. Hamidi et al. (2023) [48] also evaluated the performance of ChatGPT, Google Bard, and Claude for EHR QA. Lehman et al. (2023) [73] is another study introduced in the year 2023 that evaluated different language models (T5-Base, Clinical-T5-Base-Ckpt, Clinical-T5-Base, RoBERTa-Large, BioClinRoBERTa, GatorTron, T5-Large, Clinical-T5-Large, PubMedGPT, T5-XL, Flan-T5-XXL, GPT-3) for machine reading comprehension task on EHR notes.

For QA over structured EHR tables, TREQS [5], MedTS [62], and T5 [71] models are used. The TRanslate-Edit Model for Question-to-SQL (TREQS) [5] is a sequence-to-sequence model that uses a question encoder to convert the questions into vector representations, which are then decoded into SQL queries by the decoder. The generated SQL queries are further edited using an attentive-copying mechanism and recovery mechanism. The generated query's condition values are compared to the closest value in the database, and the condition value is replaced by this value in the database. Medical text-to-SQL model (MedTS) [62] is another text-to-SQL model that uses a pre-trained BERT model as an encoder and a grammar-based LSTM decoder to obtain an intermediate sequence. Experiments on the MIMICSQL dataset have shown that the MedTS model outperforms the TREQS model by 41% logical form accuracy and by 37% execution accuracy. Note that logical form accuracy and execution accuracy are some common evaluation metrics in text-to-SQL tasks. They are explained in detail in the subsection Evaluation Metrics. Some other examples of table-based QA methods include Tranx [82], Coarse2Fine [83], transformer-based model [63], lexicon-based models [63], quEHRy [72], and sequence-to-sequence task used with ParaGen and ParaDetect models [8].

Some models for QA over graph-based EHR are the sequence-to-sequence model [41], TREQS model [41], UniQA model [61], and attention-based aspect reasoning (AAR) method for KBQA [40]. For majority of these models [41, 61], the EHR relational database (like MIMIC-III) is converted into a knowledge graph and a question-to-SPARQL task is performed in order to retrieve answers from the knowledge graph. The sequence-to sequence model [81] uses a bidirectional LSTM as the encoder and uses an LSTM decoder while having an attention paradigm. Unlike the TREQS model [5], the sequence-to-sequence model cannot handle out-of-vocabulary words. The UniQA model [61] uses unified encoder-decoder architecture along with input-masking (IM) and value recovering technique, thus it is robust to typos and mistakes in questions. The UniQA model can be used for both table-based QA datasets (MIMICSQL*) and graph-based QA datasets (MIMIC-SPARQL*). The condition value of the query generated using the question-to-query model is compared with the values in the database. This is called the condition value recovery technique. ROUGE-L score [84] is used to check the similarity between the values in the database to that of the condition values in the generated query. Then, the condition values are replaced with values most similar to those in the database. After applying the recovery technique, UniQA outperforms both the sequence-to-sequence model (by 301.35% logical form accuracy and 170.23% execution accuracy) and the TREQS model (by 46.06% logical form accuracy and 30.76% execution accuracy). Wang et al. (2021) [40] developed a knowledge base by linking the clinical notes of patients and introduced an attention-based aspect reasoning (AAR) method for KBQA.

Most of the existing works discuss only QA on unimodal EHR data. Bardhan et al. (2022) [23] has proposed a simple pipeline for multimodal QA on EHRs (called MultimodalEHRQA) that uses a modality selection network in order to choose the modality between structured and unstructured EHR as the preferred modality. If the selected modality obtained is "unstructured text", then QA is performed over the clinical notes using BERT or ClinicalBERT, and the span of text from the clinical notes is returned as the multimodal answer. Similarly, if the preferred modality selected is "structured tables", then a text-to-SQL task is performed using the TREQS model [5]. Further research is still needed to develop a multimodal QA model capable of handling the more challenging task of using answers from both structured and unstructured data to obtain a contextualized answer.

**Evaluation Metrics**
In this section, we discuss the different evaluation metrics used for EHR. Evaluation metrics are used to evaluate the efficacy of different models. Multimedia Appendix 4 lists the different evaluation metrics used in different EHR QA studies.

The type of QA task would determine the evaluation metrics used. For QA with machine reading comprehension tasks (for example, in QA over clinical notes), exact match and F1 score are the most popular metric for evaluation [24, 25, 42, 46, 50, 51, 60, 65, 66, 68]. Exact match refers to the percentage of predictions that exactly match the ground truth answers. In [24], an exact match is used to determine if the answer entity is included in the evidence. If not, it is determined whether the projected span of evidence is within a few characters of the actual evidence. The F1 measure is a broader metric that calculates the average overlap between the prediction and the correct answer [6]. It is defined as the harmonic mean of precision and recall. Wen et al. (2020) [64] and Moon et al. (2022) [67] used exact match and partial match to assess the QA models for answering questions based on patient-specific clinical text. F1 measure was used for weighing the partial match between the predicted token of words and the golden token of words.

Evaluation metrics such as logical form accuracy and execution accuracy are commonly used for evaluating models responsible for table-based QA that use a question-to-SQL query-based approach [5, 62, 71]. They are also used for graph-based QA that use a question-to-SPARQL query-based approach [41, 61]. The logical form accuracy is calculated by doing a string comparison between the predicted SQL/SPARQL queries and the ground truth queries, and execution accuracy is calculated by obtaining the ratio of the number of generated queries that produce correct answers to the total number of queries [5]. There are instances where execution accuracy might include questions where the generated SQL query is different from the ground truth query, but the returned answer is the same. Structural accuracy is another metric used to evaluate models used for question-to-SQL/question-to-SPARQL query tasks [41, 61]. Structural accuracy is similar to measuring logical form accuracy, except that it ignores the condition value tokens. Condition value refers to the string value or numeric value in the WHERE part of the SQL/SPARQL query. For example, in the SQL query - "SELECT MAX(age) from patients WHERE Gender = "F" and DoB > 2020", "F" and 2020 are the condition values. The objective of using structural accuracy is to evaluate the accuracy of converting questions to SQL/SPARQL query structures, by not giving importance to the condition values (like in Spider dataset [85]). Raghavan et al. (2021) [8] uses exact match and denotation accuracy for evaluating clinical table-QA models. The framework involves two stages - (1) Predicting logical forms for questions, and (2) obtaining answers from the database with logical forms as input. Exact match is used for semantic parsing, while denotation accuracy is used to evaluate models for obtaining answers from logical forms. Denotation accuracy checks if the logical forms which are input to the model return the correct label answer, and the exact match is used to check if the logical forms generated are the same as the ground truth logical forms.

A variety of text-generating metrics have been used to evaluate question paraphrasing. Soni et al. (2019) [52] used BLEU [86], METEOR [87], and TER [88] for evaluating paraphrasing models. The BLEU (or Bilingual Evaluation Understudy) score evaluates how closely generated paraphrases (or candidate translation) resemble those in the reference. This is done with exact token matching. The METEOR score (Metric for Evaluation of Translation with Explicit ORdering) on the other hand uses synonyms and word stems. The edit distance (the number of edits necessary to change one sentence into another) between generated and reference paraphrases is measured by the TER score (Translation Error Rate). It is calculated by adding up all the edits, dividing that total by the number of words, and multiplying that result by 100.

## Discussion

### Challenges and existing solutions

#### *Limited number of clinical annotations for constructing EHR QA datasets*
There are very few clinical EHR annotations that are publicly available. n2c2 repository is one of the very few public repositories that hosts EHR NLP datasets (that can be used to create template-based QA datasets). This is because creating these annotations requires a lot of manual work which can be time-consuming, and at the same time require domain knowledge [23, 24]. For the same reasons, it was difficult to annotate EHR QA datasets. There are also some ethical issues and privacy concerns which need to be handled while constructing EHR QA datasets. This involves the de-identification of information related to patients.

Datasets like emrQA [24] and ClinicalKBQA [40] are examples of template-based datasets that have used the available expert annotations of the n2c2 repository to generate large-scale patient-specific QA datasets using semi-automated methods, taking advantage of the limited clinical annotations. While questions in these datasets do not represent the true distribution of questions one would ask to EHR, their scale makes them valuable for transfer learning and methods development.

#### *Concept normalization in clinical QA*
Question answering in any domain has its own challenges. But clinical QA has added challenges. One major challenge is when different phrases are used for the same medical concept in the question and the database. To deal with this issue, clinical normalization is used. Clinical normalization involves recognizing the medical entities and terminologies and converting them into a singular clinical terminology or language. A lot of EHR QA datasets such as emrQA have used MetaMap [89] during the dataset generation process to map medical terminologies mentioned in the clinical text to the UMLS Metathesaurus. However, it has been argued that concept normalization for EHR QA is fundamentally different than the task on clinical notes [72], so QA-specific datasets are clearly needed.

#### *Generating realistic EHR QA datasets*
It is necessary to make sure that questions in EHR QA datasets contain realistic questions that clinicians and patients would want answered from EHR data. In order to create realistic questions while constructing the EHRSQL dataset [36], a poll was created at a hospital to gather real-world questions that are frequently asked on the structured EHR data. The DiSCQ dataset [43] also included clinically relevant questions by collecting questions that physicians could ask. This ensured the use of medically relevant questions in the EHR QA datasets.

Adding more paraphrases to the QA dataset is another way to make sure the questions are realistic. This is because, in a real-world scenario, the same question may be posed or stated in different ways. Generation of paraphrases may be machine-generated, human-generated [24], or it could be a combination of both [36]. Table 5 lists the number of paraphrases used per template in different EHR QA datasets.

Table 5. Paraphrasing of questions in EHR QA datasets

| Dataset | Average no. of paraphrases per question type | Method of generating paraphrases | Total number of questions |
|---|---|---|---|
| MIMICSQL [5] | 1 | Human labor (crowdsourcing) | 10,000 questions |
| emrQA [24] | 7 | Human labor (Templates generated by physician were slot-filled) | 1 million questions |
| emrKBQA [8] | 7.5 | Human labor (Templates generated by physicians were slot-filled) | 940,173 |
| EHRSQL [36] | 21 | Human labor and machine learning | 24,000 |

**Open Issues and Future Work**

*Redundancy in the types of clinical questions*
Most of the existing EHR QA datasets are template-based datasets that are obtained by slot-filling. These datasets have several instances of the same type of templates that are slot-filled with various entities. As a result, there is redundancy in the diversity of questions generated. This is still an ongoing issue that needs to be addressed.

*Need for multimodal EHR QA systems*
Clinical EHRs contain a vast amount of patient information. They contain information about the patient's medical history, diagnosis information, discharge information, information related to medicine dosage, as well as medication allergies. Structured EHR data contains highly complementary data that may or may not be present in the clinical notes. The information in structured and unstructured EHR data may contain information that is similar, may contradict, or can provide additional context between these sources. There is a clear need for EHR QA systems that reason across both types of data.

DrugEHRQA [23] and MedAlign [49] datasets are the only multimodal EHR QA datasets that uses both structured data and unstructured clinical notes to answer questions (though MedAlign dataset is technically a pseudo-multimodal EHR QA dataset since the QA pairs of the MedAlign dataset are based on an XML markup that are derived from structured and unstructured EHR data). Bardhan et al. (2022) [23] introduced a simple baseline QA model for multimodal EHR data, and further research is needed to develop a multimodal QA model that unifies the EHR data modalities to obtain a contextualized answer.

### *QA of EHRs on unseen paraphrased questions*
QA models trained on clinical question-answer pairs when tested on unseen paraphrased questions have historically produced poor results. There have been works that have tried to address this challenge. The model in Raghavan et al. (2021) [8] uses paraphrasing detection and generation as a supplementary task to handle this issue. Another solution was discussed in Rawat et al. (2020) [60]. Rawat et al. (2020) [60] introduced a multi-task learning approach where extractive QA and prediction of answer span was the primary task, with an auxiliary task of logical form prediction for the questions. But this is still an ongoing issue which needs further work.

### *QA of EHRs on unseen data*
QA models should be able to generalize to new clinical contexts and EHR questions. In order to study generalization, Yue et al. (2020) [46] evaluated the performance of a model trained on the emrQA dataset on a new set of questions based on clinical notes of MIMIC-III. The experiment proved that the accuracy of the QA model dropped down by 40% when tested on unseen data. The same research group later proposed a solution [42]. They developed the CliniQG4QA framework, which uses question generation to obtain QA pairs for unseen clinical notes and strengthen QA models without the need for manual annotations. This was done using a sequence-to-sequence-based question phrase prediction model (QPP).

This issue was also addressed in question-to-SQL tasks for table-based EHR QA. Tarbell et al. (2023) [71] introduced the MIMICSQL 2.0 data split (derived from the existing MIMICSQL dataset [5]) to test generalizability of existing text-to-SQL models on EHRs. The performance of TREQS [5] model on the MIMICSQL 2.0 data split was drastically poor (logical form accuracy of 0.068 and execution accuracy of 0.173 when trained on paraphrased questions and tested on paraphrased questions), thus showing the need for improvement. To improve generalizability of text-to-SQL tasks on EHR data, Tarbell et al. (2023) then introduced the use of T5 model with data augmentation method using back-translation and further adding out-of-domain training data to improve generalizability on text-to-SQL tasks. The proposed model even though outperformed the TREQS model (logical form accuracy of 0.233 and execution accuracy of 0.528 when trained on paraphrased questions and tested on paraphrased questions), still needs further improvement. More work is required in the future to overcome this challenge.

### **Strengths**
In this study, we presented the first scoping review for QA on EHRs. We methodologically collected and screened papers related to EHR QA from January 1st,

2005 to September 30th, 2023, and performed a thorough review of the existing studies on EHR question answering. Then, we explored all the existing datasets, approaches, and evaluation metrics used in EHR QA. Furthermore, we identified the different modalities for QA over EHRs and described the approaches used for each. We have fulfilled all PRISMA scoping review requirements.

This review helps to identify the challenges faced in EHR QA. Also, this study sheds light on the problems that have been solved along with the additional gaps that are still remaining. This will encourage researchers in this domain to pursue these open problems that have not yet been solved.

**Limitations**

Despite the strengths of this study, we note a few limitations. First, the search process was limited to a handful of EHR and QA-related keywords. There is a long tail in how these types of systems are described in the literature, but there is a possibility that we might have missed relevant studies that did not match this initial search criteria. We used forward snowballing to partially resolve this issue. This helped us to identify ten additional papers that we had missed out on earlier. But in spite of this, there is still a slim chance that we might have missed a few relevant studies in our final list. Furthermore, given the current expansion of research into EHR QA, we predict that new studies would been added to this list since our search.

**Conclusions**

In recent years, question answering over EHRs has made significant progress. This is the first systematic/scoping review of QA over EHRs. In this paper, we have provided a detailed review of the different approaches and techniques used for EHR QA. The study began by discussing the need for large domain specific EHR QA datasets and then discussed the existing EHR QA datasets. We have reviewed the different unimodal EHR QA models used for both structured EHRs and unstructured EHRs, as well as QA models on multimodal EHRs. Then, we identified the major challenges in this field, such as the limited number of clinical annotations available for EHR QA dataset generation. We also talked about potential future directions in this field. It is a relatively new field with many unexplored challenges that require attention. This study should help future researchers to explore various research directions within EHR QA and expand the horizons of research areas in this field.

**Acknowledgements**

This project has been funded by NIH grants R00LM012104, R21EB029575, and R01LM011934.

**Author's Contributions**

For this study, JB, KR, and DZW proposed the idea of the study. All the authors (JB, KR, DZW) jointly made the rules for inclusion and exclusion criteria. JB and KR contributed towards paper collection and overall screening process. Both JB and KR classified the papers based on their scope. JB conducted the initial analysis and drafted the manuscript. The manuscript of the paper was then critically reviewed by KR and DZW. All authors approved the final version of the manuscript.

**Conflicts of Interest**
None declared.

**Abbreviations**
EHR: Electronic Health Record
EMR: Electronic Medical Record
KG: Knowledge Graph
KB: Knowledge Base
NLP: Natural Language Processing
QA: Question Answering

# References

1. Demner-Fushman D, Mrabet Y, Ben Abacha A. Consumer health information and question answering: helping consumers find answers to their health-related information needs. *Journal of the American Medical Informatics Association*. 2020;27(2):194-201 doi: 10.1093/jamia/ocz152.
2. Roberts K, Masterton K, Fiszman M, Kilicoglu H, Demner-Fushman D. Annotating Question Decomposition on Complex Medical Questions. Paper presented at: LREC, 2014 http://www.lrec-conf.org/proceedings/lrec2014/pdf/124_Paper.pdf.
3. Cairns BL, Nielsen RD, Masanz JJ, et al. The MiPACQ clinical question answering system. Paper presented at: AMIA annual symposium proceedings, 2011 PMID: 22195068.
4. Lee M, Cimino J, Zhu HR, et al. Beyond information retrieval—medical question answering. Paper presented at: American Medical Informatics Association, 2006 PMID: 17238385 https://www.ncbi.nlm.nih.gov/pmc/articles/PMC1839371/.
5. Wang P, Shi T, Reddy CK. Text-to-SQL generation for question answering on electronic medical records. Paper presented at: Proceedings of The Web Conference 2020, 2020 doi:10.1145/3366423.3380120.
6. Mutabazi E, Ni J, Tang G, Cao W. A review on medical textual question answering systems based on deep learning approaches. *Applied Sciences*. 2021;11(12):5456 doi: 10.3390/app11125456.
7. Athenikos SJ, Han H. Biomedical question answering: A survey. *Computer methods and programs in biomedicine*;99(1):1-24 PMID: 19913938 doi: 10.1016/j.cmpb.2009.10.003.
8. Raghavan P, Liang JJ, Mahajan D, Chandra R, Szolovits P. emrkbqa: A clinical knowledge-base question answering dataset. Paper presented at: Proceedings of the 20th Workshop on Biomedical Language Processing, 2021 doi: 10.18653/v1/2021.bionlp-1.7 https://aclanthology.org/2021.bionlp-1.7.
9. Johnson AE, Bulgarelli L, Shen L, et al. MIMIC-IV, a freely accessible electronic health record dataset. *Scientific data*. 2023;10(1):1 PMID: 36596836 doi:10.1038/s41597-022-01899-x.

10. Pollard TJ, Johnson AE, Raffa JD, Celi LA, Mark RG, Badawi O. The eICU Collaborative Research Database, a freely available multi-center database for critical care research. *Scientific data*. 2018;5(1):1-13 PMID: 30204154 doi: 10.1038/sdata.2018.178.
11. Johnson AE, Pollard TJ, Shen L, et al. MIMIC-III, a freely accessible critical care database. *Scientific data*. 2016;3(1):1-9 PMID: 27219127 doi: 10.1038/sdata.2016.35.
12. Datta S, Roberts K. Fine-grained spatial information extraction in radiology as two-turn question answering. *International journal of medical informatics*. 2022;158:104628 PMID: 34839119 doi: 10.1016/j.ijmedinf.2021.104628.
13. Xiong Y, Peng W, Chen Q, Huang Z, Tang B. A Unified Machine Reading Comprehension Framework for Cohort Selection. *IEEE Journal of Biomedical and Health Informatics*. 2021;26(1):379-387 PMID: 34236972 doi: 10.1109/JBHI.2021.3095478.
14. Devlin J, Chang MW, Lee K, Toutanova K. Bert: Pre-training of deep bidirectional transformers for language understanding. Paper presented at: Proceedings of NAACL-HLT 2019, 2019 doi: 10.48550/arXiv.1810.04805 https://aclanthology.org/N19-1423.pdf.
15. Seo M, Kembhavi A, Farhadi A, Hajishirzi H. Bidirectional attention flow for machine comprehension. Paper presented at: 5th International Conference on Learning Representations, 2017 doi: 10.48550/arXiv.1611.01603.
16. Lee J, Yoon W, Kim S, et al. BioBERT: a pre-trained biomedical language representation model for biomedical text mining. *Bioinformatics*. 2020;36(4):1234-1240 PMID: 31501885 doi: 10.1093/bioinformatics/btz682.
17. Zhuang L, Wayne L, Ya S, Jun Z. A Robustly Optimized BERT Pre-training Approach with Post-training. Paper presented at: Proceedings of the 20th Chinese National Conference on Computational Linguistics, 2021 https://aclanthology.org/2021.ccl-1.108.
18. Wang Z, Hamza W, Florian R. Bilateral multi-perspective matching for natural language sentences. Paper presented at: Proceedings of the Twenty-Sixth International Joint Conference on Artificial Intelligence (IJCAI-17), 2017 doi: 10.48550/arXiv.1702.03814.
19. Liang JJ, Lehman E, Iyengar A, et al. Towards Generalizable Methods for Automating Risk Score Calculation. Paper presented at: Proceedings of the 21st Workshop on Biomedical Language Processing, 2022 doi: 10.18653/v1/2022.bionlp-1.42 URL: https://aclanthology.org/2022.bionlp-1.42.
20. Newman-Griffis D, Divita G, Desmet B, Zirikly A, Rose CP, Fosler-Lussier E. Ambiguity in medical concept normalization: An analysis of types and coverage in electronic health record datasets. *Journal of the American Medical Informatics Association*. 2021 PMID: 33319905 doi: 10.1093/jamia/ocaa269;28(3):516-532.

21. Wells BJ, Chagin KM, Nowacki AS, Kattan MW. Strategies for handling missing data in electronic health record derived data. *EGEMS*. 2013 PMID: 25848578 doi: 10.13063/2327-9214.1035;1(3).
22. Haneuse S, Arterburn D, Daniels MJ. Assessing missing data assumptions in EHR-based studies: a complex and underappreciated task. *JAMA Network Open*. 2021;4(2):e210184--e210184 PMID: 33635321 doi: 10.1001/jamanetworkopen.2021.0184.
23. Bardhan J, Colas A, Roberts K, Wang DZ. Drugehrqa: A question answering dataset on structured and unstructured electronic health records for medicine related queries. Paper presented at: Proceedings of the 13th Conference on Language Resources and Evaluation (LREC 2022), 2022 doi: 10.48550/arXiv.2205.01290 https://aclanthology.org/2022.lrec-1.117.
24. Pampari A, Raghavan P, Liang J, Peng J. emrqa: A large corpus for question answering on electronic medical records. Paper presented at: Proceedings of the 2018 Conference on Empirical Methods in Natural Language Processing, 2018 doi: 10.18653/v1/D18-1258 https://aclanthology.org/D18-1258.
25. Moon S, He H, Jia H, Liu H, Fan JW. Extractive Clinical Question-Answering With Multianswer and Multifocus Questions: Data Set Development and Evaluation Study. *JMIR AI*. 2023 doi:10.2196/41818;2(1):e41818.
26. Uzuner O, Solti I, Xia F, Cadag E. Community annotation experiment for ground truth generation for the i2b2 medication challenge. *Journal of the American Medical Informatics Association*. 2010;17(5):519-523 PMID: 20819855 doi: 10.1136/jamia.2010.004200.
27. Uzuner O, Solti I, Cadag E. Extracting medication information from clinical text. *Journal of the American Medical Informatics Association*. 2010;17(5):514-518 PMID: 20819854 doi: 10.1136/jamia.2010.003947.
28. Uzuner O, Goldstein I, Luo Y, Kohane I. Identifying patient smoking status from medical discharge records. *Journal of the American Medical Informatics Association*. 2008;15(1):14-24 PMID: 17947624 doi: 10.1197/jamia.M2408.
29. Uzuner O. Recognizing obesity and comorbidities in sparse data. *Journal of the American Medical Informatics Association*. 2009;16(4):561-570 PMID: 19390096 doi: 10.1197/jamia.M3115.
30. Stubbs A, Uzuner. Annotating longitudinal clinical narratives for de-identification: The 2014 i2b2/UTHealth corpus. *Journal of biomedical informatics*. 2015;58:S20-S29 PMID: 26319540 doi: 10.1016/j.jbi.2015.07.020.
31. Uzuner O, Bodnari A, Shen S, Forbush T, Pestian J, South BR. Evaluating the state of the art in coreference resolution for electronic medical records. *Journal of the American Medical Informatics Association*. 2012;19(5):786-791 PMID: 22366294 doi: 10.1136/amiajnl-2011-000784.
32. Roberts K, Demner-Fushman D. Annotating logical forms for EHR questions. Paper presented at: Proceedings of the Tenth International Conference on Language Resources and Evaluation (LREC'16), 2016 PMID: 28503677 https://aclanthology.org/L16-1598.

33. Roberts K, Demner-Fushman D. Toward a natural language interface for EHR questions. Paper presented at: AMIA Summits on Translational Science Proceedings, 2015 PMID: 26306260 https://www.ncbi.nlm.nih.gov/pmc/articles/PMC4525248/.
34. Raghavan P, Patwardhan S, Liang JJ, Devarakonda MV. Annotating Electronic Medical Records for Question Answering. *arXiv Preprint*. May 17, 2018 doi: 10.48550/arXiv.1805.06816 https://arxiv.org/ftp/arxiv/papers/1805/1805.06816.pdf.
35. Fan J. Annotating and Characterizing Clinical Sentences with Explicit Why-QA Cues. Paper presented at: Proceedings of the 2nd Clinical Natural Language Processing Workshop, 2019 doi: 10.18653/v1/W19-1913 https://aclanthology.org/W19-1913.
36. Lee G, Hwang H, Bae S, et al. EHRSQL: A Practical Text-to-SQL Benchmark for Electronic Health Records. Paper presented at: 36th Conference on Neural Information Processing Systems (NeurIPS 2022) Track on Datasets and Benchmarks, 2022 https://proceedings.neurips.cc/paper_files/paper/2022/file/643e347250cf9289e5a2a6c1ed5ee42e-Paper-Datasets_and_Benchmarks.pdf.
37. Soni S, Gudala M, Pajouhi A, Roberts K. RadQA: A Question Answering Dataset to Improve Comprehension of Radiology Reports. Paper presented at: Proceedings of the Thirteenth Language Resources and Evaluation Conference, 2022 https://aclanthology.org/2022.lrec-1.672.
38. Oliveira LESe, Schneider ETR, Gumiel YB, Luz MAPd, Paraiso EC, Moro C. Experiments on Portuguese clinical question answering. Paper presented at: Intelligent Systems: 10th Brazilian Conference, BRACIS 2021, 2021 doi: 10.1007/978-3-030-91699-2_10.
39. Kim D, Bae S, Kim S, Choi E. Uncertainty-Aware Text-to-Program for Question Answering on Structured Electronic Health Records. Paper presented at: Proceedings of Machine Learning Research, Conference on Health, Inference, and Learning, 2022 https://proceedings.mlr.press/v174/kim22a/kim22a.pdf.
40. Wang P, Shi T, Agarwal K, Choudhury S, Reddy CK. Attention-based aspect reasoning for knowledge base question answering on clinical notes. Paper presented at: Proceedings of the 13th ACM International Conference on Bioinformatics, Computational Biology and Health Informatics, 2022 doi: 10.1145/3535508.3545518.
41. Park J, Cho Y, Lee H, Choo J, Choi E. Knowledge graph-based question answering with electronic health records. Paper presented at: Proceedings of Machine Learning Research, Machine Learning for Healthcare Conference, 2021 https://proceedings.mlr.press/v149/park21a/park21a.pdf.
42. Yue X, Zhang XF, Yao Z, Lin S, Sun H. Cliniqg4qa: Generating diverse questions for domain adaptation of clinical question answering. Paper presented at: 2021 IEEE International Conference on Bioinformatics and Biomedicine

(BIBM), 2021 doi: 10.1109/bibm52615.2021.9669300 https://ieeexplore.ieee.org/document/9669300.

43. Lehman E, Lialin V, Legaspi KY, et al. Learning to Ask Like a Physician. Paper presented at: Proceedings of the 4th Clinical Natural Language Processing Workshop, 2022 doi: 10.18653/v1/2022.clinicalnlp-1.8 https://aclanthology.org/2022.clinicalnlp-1.8.

44. Soni S, Gudala M, Wang DZ, Roberts K. Using FHIR to construct a corpus of clinical questions annotated with logical forms and answers. Paper presented at: AMIA Annual Symposium Proceedings, 2019 PMID: 32308918 https://pubmed.ncbi.nlm.nih.gov/32308918/.

45. Mishra S, Awasthi R, Papay F, et al. DiagnosisQA: A semi-automated pipeline for developing clinician validated diagnosis specific QA datasets. *medRxiv*. 2021 doi: 10.1101/2021.11.10.21266184 https://www.medrxiv.org/content/10.1101/2021.11.10.21266184v1.

46. Yue X, Gutierrez BJ, Sun H. Clinical reading comprehension: a thorough analysis of the emrQA dataset. Paper presented at: Proceedings of the 58th Annual Meeting of the Association for Computational Linguistics, 2020 doi: 10.18653/v1/2020.acl-main.410 https://aclanthology.org/2020.acl-main.410.

47. Pal A. CLIFT: Analysing Natural Distribution Shift on Question Answering Models in Clinical Domain. Paper presented at: NeurIPS 2022 Workshop on Robustness in Sequence Modeling, 2022 https://openreview.net/pdf?id=9PQFROOfqm.

48. Hamidi A, Roberts K. Evaluation of AI Chatbots for Patient-Specific EHR Questions. Paper presented at: arXiv preprint arXiv:2306.02549, 2023 DOI: 10.48550/arXiv.2306.02549.

49. Fleming SL, Lozano A, Haberkorn WJ, et al. MedAlign: A Clinician-Generated Dataset for Instruction. Paper presented at: arXiv preprint, 2023 DOI: 10.48550/arXiv.2308.14089.

50. Mahbub M, Goethert I, Danciu I, et al. Question-Answering System Extracts Information on Injection Drug Use from Clinical Progress Notes. Paper presented at: arXiv preprint, 2023 DOI: 10.48550/arXiv.2305.08777.

51. Dada A, Ufer TL, Kim M, et al. Information extraction from weakly structured radiological reports. *European Radiology*. 2023 PMID: 37505252 DOI: 10.1007/s00330-023-09977-3:1-8.

52. Soni S, Roberts K. A paraphrase generation system for ehr question answering. Paper presented at: Proceedings of the 18th BioNLP Workshop and Shared Task, 2019 doi: 10.18653/v1/W19-5003 https://aclanthology.org/W19-5003.

53. Moon SR, Fan J. How you ask matters: The effect of paraphrastic questions to bert performance on a clinical squad dataset. Paper presented at: Proceedings of the 3rd Clinical Natural Language Processing Workshop, 2020 doi: 10.18653/v1/2020.clinicalnlp-1.13 https://aclanthology.org/2020.clinicalnlp-1.13.

54. Soni S, Roberts K. Paraphrasing to improve the performance of electronic health records question answering. Paper presented at: AMIA Summits on Translational Science Proceedings, 2020 PMID: 32477685 https://pubmed.ncbi.nlm.nih.gov/32477685/.
55. Patrick J, Li M. An ontology for clinical questions about the contents of patient notes. *Journal of Biomedical Informatics*. 2012;45(2):292-306 PMID: 22142949 doi: 10.1016/j.jbi.2011.11.008.
56. Roberts K, Rodriguez L, Shooshan SE, Demner-Fushman D. Resource classification for medical questions. Paper presented at: AMIA Annual Symposium Proceedings, 2016 PMID: 28269901 https://www.ncbi.nlm.nih.gov/pmc/articles/PMC5333297/.
57. Roberts K, Patra BG. A semantic parsing method for mapping clinical questions to logical forms. Paper presented at: AMIA Annual Symposium Proceedings, 2017 PMID: 29854217 https://pubmed.ncbi.nlm.nih.gov/29854217/.
58. Schwertner MA, Rigo SJ, Araujo DA, Silva AB, Eskofier B. Fostering natural language question answering over knowledge bases in oncology EHR. Paper presented at: 2019 IEEE 32nd International Symposium on Computer-Based Medical Systems (CBMS), 2019 doi: 10.1109/CBMS.2019.00102 https://ieeexplore.ieee.org/document/8787420.
59. Rawat BPS, Li FaYH. Clinical Judgement Study using Question Answering from Electronic Health Records. Paper presented at: Machine Learning for Healthcare Conference, 2019 PMID: 31897452 https://pubmed.ncbi.nlm.nih.gov/31897452/.
60. Rawat BPS, Weng WH, Min SY, Raghavan P, Szolovits P. Entity-enriched neural models for clinical question answering. Paper presented at: Proceedings of the BioNLP 2020 workshop, 2020 doi: 10.18653/v1/2020.bionlp-1.12 https://aclanthology.org/2020.bionlp-1.12.pdf.
61. Bae S, Kim D, Kim J, Choi E. Question answering for complex electronic health records database using unified encoder-decoder architecture. Paper presented at: Proceedings of Machine Learning Research, Machine Learning for Health, 2021 https://arxiv.org/abs/2111.14703.
62. Pan Y, Wang C, Hu B, et al. A BERT-Based Generation Model to Transform Medical Texts to SQL Queries for Electronic Medical Records: Model Development and Validation. *JMIR Medical Informatics*. 2021;9(12):e32698 PMID: 34889749 doi: 10.2196/32698.
63. Soni S, Roberts K. Toward a Neural Semantic Parsing System for EHR Question Answering. Paper presented at: AMIA Annual Symposium Proceedings, 2022 PMID: 37128406 https://pubmed.ncbi.nlm.nih.gov/37128406/.
64. Wen A, Elwazir MY, Moon S, Fan J. Adapting and evaluating a deep learning language model for clinical why-question answering. *JAMIA open*. 2020;3(1):16-20 PMID: 32607483 doi: 10.1093/jamiaopen/ooz072.
65. Soni S, Roberts K. Evaluation of dataset selection for pre-training and fine-tuning transformer language models for clinical question answering. Paper presented at: Proceedings of the Twelfth Language Resources and Evaluation Conference, 2020 https://aclanthology.org/2020.lrec-1.679.

66. Mairittha T, Mairittha N, Inoue S. Improving fine-tuned question answering models for electronic health records. Paper presented at: Adjunct Proceedings of the 2020 ACM international joint conference on pervasive and ubiquitous computing and proceedings of the 2020 ACM International Symposium on Wearable Computers, 2020 doi: 10.1145/3410530.3414436.

67. Moon S, He H, Fan JW. Effects of Information Masking in the Task-Specific Finetuning of a Transformers-Based Clinical Question-Answering Framework. Paper presented at: 2022 IEEE 10th International Conference on Healthcare Informatics (ICHI), 2022 doi: 10.1109/ichi54592.2022.00017 https://ieeexplore.ieee.org/document/9874588.

68. Li Y, Wehbe RM, Ahmad FS, Wang H, Luo Y. A comparative study of pretrained language models for long clinical text. *Journal of the American Medical Informatics Association*. 2023;30(2):340-347 PMID: 36451266 doi: 10.1093/jamia/ocac225.

69. Yang X, Chen A, PourNejatian N, et al. A large language model for electronic health records. *npj Digital Medicine*. 2022;5(1):194 PMID: 36572766 doi: 10.1038/s41746-022-00742-2.

70. Kang M, Baek J, Hwang SJ. KALA: Knowledge-Augmented Language Model Adaptation, 2022 DOI: 10.48550/arXiv.2204.10555.

71. Tarbell R, Choo KKR, Dietrich GaRA. Towards Understanding the Generalization of Medical Text-to-SQL. Paper presented at: arXiv preprint, 2023 DOI: 10.48550/arXiv.2303.12898.

72. Soni S, Datta S, Roberts K. quEHRy: a question answering system to query electronic health records. *Journal of the American Medical Informatics Association PMID: 37087111 DOI: 10.1093/jamia/ocad050*. 2023;30(6):1091-1102.

73. Lehman E, Hernandez E, Mahajan D, et al. Do We Still Need Clinical Language Models? Paper presented at: Proceedings of Machine Learning Research, 2023 DOI: 10.48550/arXiv.2302.08091.

74. Tsui FR, Shi L, Ruiz V, et al. Natural language processing and machine learning of electronic health records for prediction of first-time suicide attempts. *JAMIA open PMID: 33758800 doi: 10.1093/jamiaopen/ooab011*. 2021;4(1).

75. Yue X, Zhang XF, Sun H. Annotated Question-Answer Pairs for Clinical Notes in the MIMIC-III Database. *https://physionet.org/*. 2021 https://doi.org/10.13026/j0y6-bw05.

76. Rajpurkar P, Zhang J, Lopyrev K, Liang P. Squad: 100,000+ questions for machine comprehension of text. Paper presented at: Proceedings of the 2016 Conference on Empirical Methods in Natural Language Processing, 2016 doi: 10.18653/v1/D16-1264 https://aclanthology.org/D16-1264.

77. Roberts K, Demner-Fushman D, Voorhees EM, Hersh WR. Proceedings of The Twenty-Fifth Text REtrieval Conference, TREC 2016. Paper presented at:

Nguyen, Gia-Hung and Soulier, Laure and Tamine, Lynda and Bricon-Souf, Nathalie, 2016 URL: http://trec.nist.gov/pubs/trec25/papers/Overview-CL.pdf.

78. Zhang Z, Han X, Liu Z, Jiang X, Sun M, Liu Q. ERNIE: Enhanced language representation with informative entities. Paper presented at: Proceedings of the 57th Annual Meeting of the Association for Computational Linguistics, 2019 doi: 10.18653/v1/P19-1139 https://aclanthology.org/P19-1139.
79. Suster SaDW. CliCR: a dataset of clinical case reports for machine reading comprehension. Paper presented at: Proceedings of the 2018 Conference of the North American Chapter of the Association for Computational Linguistics, 2018 doi: 10.18653/v1/N18-1140.
80. Min SY, Raghavan P, Szolovits P. Advancing seq2seq with joint paraphrase learning. Paper presented at: Proceedings of the 3rd Clinical Natural Language Processing Workshop, 2020 doi: 10.18653/v1/2020.clinicalnlp-1.30 https://aclanthology.org/2020.clinicalnlp-1.30.
81. Luong MT, Pham H, Manning CD. Effective approaches to attention-based neural machine translation. Paper presented at: Proceedings of the 2015 Conference on Empirical Methods in Natural Language Processing, 2015 doi: 10.18653/v1/D15-1166 https://aclanthology.org/D15-1166.
82. Yin P, Neubig G. Tranx: A transition-based neural abstract syntax parser for semantic parsing and code generation. Paper presented at: Proceedings of the 2018 Conference on Empirical Methods in Natural Language Processing: System Demonstrations, 2018 DOI: 10.18653/v1/D18-2002 URL: https://aclanthology.org/D18-2002.
83. Dong L, Lapata M. Coarse-to-fine decoding for neural semantic parsing. Paper presented at: Proceedings of the 56th Annual Meeting of the Association for Computational Linguistics (Volume 1: Long Papers), 2018 DOI: 10.18653/v1/P18-1068 URL: https://aclanthology.org/P18-1068.
84. Lin CY. Rouge: A package for automatic evaluation of summaries. Paper presented at: Text summarization branches out, Association for Computational Linguistics, 2004 https://aclanthology.org/W04-1013.
85. Yu T, Zhang R, Yang K, et al. Spider: A large-scale human-labeled dataset for complex and cross-domain semantic parsing and text-to-sql task. Paper presented at: Proceedings of the 2018 Conference on Empirical Methods in Natural Language Processing, 2018 doi: 10.18653/v1/D18-1425 https://aclanthology.org/D18-1425.
86. Papineni K, Roukos S, Ward T, Zhu WJ. Bleu: a method for automatic evaluation of machine translation. Paper presented at: Proceedings of the 40th annual meeting of the Association for Computational Linguistics, 2002 doi: 10.3115/1073083.1073135 https://aclanthology.org/P02-1040.pdf.
87. Agarwal A, Lavie A. METEOR: An automatic metric for MT evaluation with high levels of correlation with human judgments. Paper presented at: Proceedings of the Second Workshop on Statistical Machine Translation, 2007 https://aclanthology.org/W07-0734.
88. Snover M, Dorr B, Schwartz R, Micciulla L, Makhoul J. A study of translation edit rate with targeted human annotation. Paper presented at: Proceedings of the 7th Conference of the Association for Machine Translation in the Americas: Technical Papers, 2006 https://aclanthology.org/2006.amta-papers.25.
89. Aronson AR. Effective mapping of biomedical text to the UMLS Metathesaurus: the MetaMap program. Paper presented at: Proceedings of the AMIA Symposium, 2001 PMID: 11825149 https://www.ncbi.nlm.nih.gov/pmc/articles/PMC2243666/.
90. Demner-Fushman D, Chapman WW, McDonald CJ. What can natural language processing do for clinical decision support? *Journal of biomedical informatics*. 2009;42(5):760-772 PMID: 19683066 doi: 10.1016/j.jbi.2009.08.007.
91. Ely JW, Osheroff JA, Ebell MH, et al. Analysis of questions asked by family physicians regarding patient care. *Western Journal of Medicine*. 2000;172(5):315 PMID: 10435959 doi:10.1136/bmj.319.7206.358.
92. Ely JW, Osheroff JA, Gorman PN, et al. A taxonomy of generic clinical questions: classification study. *Bmj*. 2000;321(7258):429-432 PMID: 10938054 doi: 10.1136/bmj.321.7258.429.
93. Weiming W, Hu D, Feng M, Wenyin L. Automatic clinical question answering based on UMLS relations. Paper presented at: Third International Conference on Semantics, Knowledge and Grid (SKG 2007), 2007 doi: 10.1109/SKG.2007.126.
94. Huang X, Zhang J, Xu Z, Ou L, Tong J. A knowledge graph based question answering method for medical domain. *PeerJ Computer Science*. 2021;7:e667 PMID: 34604514 doi: 10.7717/peerj-cs.667.
95. Cao Y, Liu F, Simpson P, et al. AskHERMES: An online question answering system for complex clinical questions. *Journal of biomedical informatics*. 2011;44(2):277-288 doi: PMID: 21256977 doI: 10.1016/j.jbi.2011.01.004.
96. Zhang X, Wu J, He Z, Liu X, Su Y. Medical exam question answering with large-scale reading comprehension. Paper presented at: Proceedings of the AAAI conference on artificial intelligence, 2018 doi:10.1609/aaai.v32i1.11970.
97. Papanikolaou Y, Staib M, Grace JJ, Bennett F. Slot Filling for Biomedical Information Extraction. Paper presented at: Proceedings of the 21st Workshop on Biomedical Language Processing, 2022 doi:10.18653/v1/2022.bionlp-1.7.
98. Li X, Yin F, Sun Z, et al. Entity-Relation Extraction as Multi-Turn Question Answering. Paper presented at: Proceedings of the 57th Annual Meeting of the Association for Computational Linguistics, 2019 doi: 10.18653/v1/P19-1129 https://aclanthology.org/P19-1129.
99. Li M, Reddy RG, Wang Z, et al. Covid-19 claim radar: A structured claim extraction and tracking system. Paper presented at: Proceedings of the 60th Annual Meeting of the Association for Computational Linguistics: System

Demonstrations, 2022 doi: 10.18653/v1/2022.acl-demo.13 https://aclanthology.org/2022.acl-demo.13.
100. Cao J, Zhou X, Xiong W, et al. Electronic medical record entity recognition via machine reading comprehension and biaffine. *Discrete Dynamics in Nature and Society*. 2021 doi: 10.1155/2021/1640837.
101. Jacquemart P, Zweigenbaum P. *Towards a medical question-answering system: a feasibility study*. Vol 95: IOS Press; 2003 ISBN: 978-1-60750-939-4.
102. Zhang L, Yang X, Li S, Liao T, Pan G. Answering medical questions in Chinese using automatically mined knowledge and deep neural networks: an end-to-end solution. *BMC bioinformatics*. 2022;23(1):136 PMID: 35428175 doi: 10.1186/s12859-022-04658-2.
103. Sheng M, Li A, Bu Y, et al. DSQA: a domain specific QA system for smart health based on knowledge graph. *Web Information Systems and Applications*: Springer, Cham; 2020 doi: 10.1007/978-3-030-60029-7_20 ISBN: 978-3-030-60029-7.
104. REDD A, FORBUSH T, PALMER M. Recognizing questions and answers in EMR templates using natural language processing. *Integrating Information Technology and Management for Quality of Care*. 2014;202:149 PMID: 25000038 https://pubmed.ncbi.nlm.nih.gov/25000038/.
105. Demner-Fushman D. Clinical, Consumer Health, and Visual Question Answering. Paper presented at: Information Management and Big Data: 5th International Conference, 2019 doi: 10.1007/978-3-030-11680-4_1 https://link.springer.com/chapter/10.1007/978-3-030-11680-4_1.
106. Ruan T, Huang Y, Liu X, Xia Y, Gao J. QAnalysis: a question-answer driven analytic tool on knowledge graphs for leveraging electronic medical records for clinical research. *BMC medical informatics and decision making*. 2019;19:1-13 PMID: 30935389 doi: 10.1186/s12911-019-0798-8.
107. Wang Y, Tariq A, Khan F, Gichoya JW, Trivedi H, Banerjee I. Query bot for retrieving patients' clinical history: A COVID-19 use-case. *Journal of biomedical informatics*. 2021;123:103918 PMID: 34560275 doi: 10.1016/j.jbi.2021.103918.
108. Lalor JP, Wu H, Chen L, Mazor KM, Yu H. ComprehENotes, an instrument to assess patient reading comprehension of electronic health record notes: development and validation. *Journal of medical Internet research*. 2018;20(4):e139 PMID: 29695372 doi: 10.2196/jmir.9380.
109. Lalor JP, Woolf B, Yu H. Improving electronic health record note comprehension with NoteAid: randomized trial of electronic health record note comprehension interventions with crowdsourced workers. *Journal of medical Internet research*. 2019;21(1):e10793 PMID: 30664453 doi: 10.2196/10793.
110. Gao Y, Dligach D, Miller T, et al. DR. BENCH: Diagnostic Reasoning Benchmark for Clinical Natural Language Processing. *Journal of Biomedical Informatics*. 2023;138:104286 PMID: 36706848 doi: 10.1016/j.jbi.2023.104286.

111. Goyal Y, Khot T, Summers-Stay D, Batra D, Parikh D. Making the v in vqa matter: Elevating the role of image understanding in visual question answering. Paper presented at: Proceedings of the IEEE conference on computer vision and pattern recognition, 2017 doi: 10.1109/cvpr.2017.670.
112. Anderson P, He X, Buehler C, et al. Bottom-up and top-down attention for image captioning and visual question answering. Paper presented at: Proceedings of the IEEE conference on computer vision and pattern recognition, 2018 doi: 10.1109/cvpr.2018.00636.
113. Tan H, Bansal M. LXMERT: Learning Cross-Modality Encoder Representations. Paper presented at: Proceedings of the 2019 Conference on Empirical Methods in Natural Language Processing, 2019 doi: 10.18653/v1/d19-1514 https://aclanthology.org/D19-1514.
114. Zhang Y, Qian S, Fang Q, Xu C. Multi-modal knowledge-aware hierarchical attention network for explainable medical question answering. Paper presented at: Proceedings of the 27th ACM international conference on multimedia, 2019 doi: 10.1145/3343031.3351033.
115. Singh H, Nasery A, Mehta D, Agarwal A, Lamba J, Srinivasan BV. Mimoqa: Multimodal input multimodal output question answering. Paper presented at: Proceedings of the 2021 Conference of the North American Chapter of the Association for Computational Linguistics: Human Language Technologies, 2021 doi: 10.18653/v1/2021.naacl-main.418 https://aclanthology.org/2021.naacl-main.418.